\documentclass[conference]{ieeetran}
\usepackage{graphicx}
\usepackage{amsmath}
\usepackage{hyperref}
\usepackage[aboveskip=0pt,belowskip=0pt]{caption}
\usepackage{booktabs}
\usepackage{multirow}
\usepackage{cite}
\usepackage[table]{xcolor}
\usepackage{enumitem}
\usepackage{booktabs}
\usepackage{algorithm}
\usepackage{algpseudocode}
\floatname{algorithm}{Algorithm}

\newcommand{\DD}{\mathcal{D}}
\newcommand{\TT}{\mathcal{T}}
\newcommand{\mypar}[1]{\vspace{0.35\baselineskip}\noindent\textbf{#1}.\,}
\IEEEoverridecommandlockouts
\usepackage{times}

\title{Impact of domain adaptation in deep learning for medical image classifications}

\author{Yihang Wu$^{1}$, Ahmad~Chaddad$^{1,2,*}$\\
$^1$ Laboratory for AIPM, School of Artificial Intelligence, Guilin University of Electronic Technology, China\\
$^2$ Laboratory for Imagery, Vision and Artificial Intelligence, École de Technologie Supérieure, Canada\\
All authors are equally contributed, *Corresponding author (Email: ahmad8chaddad@gmail.com)%
\thanks{This research was funded by the National Natural Science Foundation of China \#82260360, the Guilin Innovation Platform and Talent Program \#20222C264164, Innovation Project of GUET Graduate Education 2025YCXS244, and the Guangxi Science and Technology Base and Talent Project (\#2022AC18004 and \#2022AC21040).}}

\begin{document}
\maketitle

\begin{abstract}
Domain adaptation (DA) is a quickly expanding area in machine learning that involves adjusting a model trained in one domain to perform well in another domain. While there have been notable progressions, the fundamental concept of numerous DA methodologies has persisted: aligning the data from various domains into a shared feature space. In this space, knowledge acquired from labeled source data can improve the model training on target data that lacks sufficient labels. In this study, we demonstrate the use of 10 deep learning models to simulate common DA techniques and explore their application in four medical image datasets. We have considered various situations such as multi-modality, noisy data, federated learning (FL), interpretability analysis, and classifier calibration. The experimental results indicate that using DA with ResNet34 in a brain tumor (BT) data set results in an enhancement of 4.7\% in model performance. Similarly, the use of DA can reduce the impact of Gaussian noise, as it provides $\sim 3\%$ accuracy increase using ResNet34 on a BT dataset. Furthermore, simply introducing DA into FL framework shows limited potential (e.g., $\sim 0.3\%$ increase in performance) for skin cancer classification. In addition, the DA method can improve the interpretability of the models using the gradcam++ technique, which offers clinical values. Calibration analysis also demonstrates that using DA provides a lower expected calibration error (ECE) value $\sim 2\%$ compared to CNN alone on a multi-modality dataset. The codes for our experiments are available at \url{https://github.com/AIPMLab/Domain_Adaptation}.


\end{abstract}


\IEEEpeerreviewmaketitle

\section{Introduction}

Rapid progress in artificial intelligence (AI) in the healthcare industry has led to significant accomplishments in various areas such as medical image classification, segmentation, and reconstruction \cite{10288131}. However, shifts in the data distribution, which can occur almost in the application, can dramatically impact the performance of machine learning (ML) models trained in a standard supervised manner \cite{chaddad2023enhancing}. For instance, in the medical domain, training a robust classification model always requires a large amount of data; otherwise, it will face an overfitting problem. However, a common issue in the medical domain is related to the large degree of variation in collection equipment from hospital to hospital, leading to huge data distribution discrepancies \cite{chaddad2023enhancing}. Furthermore, unlike natural data, the complex characteristics of medical data (e.g., magnetic resonance imaging, X-ray) face a challenge of choosing the most suitable method. Domain adaptation (DA) is associated with ML technique to reduce the differences in data distribution between different data, offers an efficient solution to the problem mentioned above \cite{10835760}. 

Let $X$ and $Y$ be two random variables, $X$ being a $D$-dimensional input with marginal distribution $p(X)$ and $Y$ the output to predict (e.g., one-hot label vector in classification) with conditional probability distribution $p(Y | X)$. We denote as $\DD=\{X,\, p(X)\}$ a domain and use $\TT=\{Y,\, p(Y | X)\}$ to represent a task where the goal is to learn $p$ with a parametric model such as a neural network (NN). Suppose that we train a model using data from a source domain (SD) $\DD_s$ for some task $\TT_s$, and to learn a similar task $\TT_t$ using target data $\DD_t$. If the two domains correspond ($\DD_s \sim \DD_t$) then the model trained in $\DD_s$ can also be used for task $\TT_t$. However, if the two domains are not similar ($\DD_s \not\sim \DD_t$), then the models trained on the SD $\DD_s$ are likely to perform poorly on a task of the target domain $\DD_t$. In this case, DA can use the related information from $\{\DD_s, \TT_s\}$ to learn $p(Y | X)$ for the target domain (TD) and the task. 

The task of DA is important because it strives to improve the adaptation of the classifier to test instances, especially when there is a change in the data distribution between the training and testing datasets \cite{10835760}. For example, in \cite{li2022cross}, they introduced an Adaptive Teacher framework for the task of detecting cross-domain objects. This framework combines a target-domain teacher model and a jointly trained cross-domain student model. Adversarial and mutual learning strategies are employed to learn domain-invariant features and transfer pseudo-labels from the teacher model to the student. Looking back at the history of DA, early methods used shallow models based on statistical learning. These methods typically follow these basic steps: 1) feature extraction, 2) matching the feature distributions of training and test sets, and 3) using a classifier for prediction. Currently, DA development is driven by advances in deep learning (DL) \cite{yang2023tvt}. Despite these advancements, a comprehensive evaluation of recent deep models with DA on medical datasets is necessary. Furthermore, previous studies have focused on improving the classification metrics of DA models, and the calibration effects of the DA model have been less explored \cite{ma2022test}.

Motivated by previous challenges, our study presents a comprehensive framework for the use of DA approaches in medical imaging. We present detailed simulations using 10 deep models with common DA techniques on four publicly available medical datasets. We also perform interpretability analysis with quantitative metrics (i.e., confidence change), and calibration analysis to validate DA models. These experiments are beneficial for understanding the impact of DA in the medical domain. In summary, the contributions of this study can be listed as follows.

\begin{itemize}
  \item We employ four publicly available medical datasets to simulate typical DA algorithms in clinical settings, covering various cases such as noisy source data, federated DA, and t-SNE visualization. Furthermore, we adopt five XAI techniques to validate the interpretability of the adaptation models.
  \item We perform calibration analysis using 10 deep models with DA on four medical datasets to examine the calibration effects of these deep models after adaptation. 
\end{itemize}

The remainder of this paper is structured as follows. Section \ref{S4} presents the common DA methods. In Section \ref{S:3}, we simulate the experiments with a thorough analysis using DA based on 10 deep CNN models in four public medical data sets. Section \ref{S8} gives the conclusion.

\section{The main method of domain adaptation in the medical fields}\label{S4}

Currently, several DA algorithms such as maximum mean discrepancy (MMD) based, adversarial (DANN) based have been applied in various medical fields \cite{guan2021domain}. In this section, DANN \cite{ganin2016domain} is analyzed as a classical example. DANN aims to learn more general features from the target domain during training, where an additional domain classifier is added to the intermediate feature layer to distinguish the domains to which different samples belong. The network consists of three modules: feature extractor, label predictor, and domain classifier ($\textbf{D}$). Let $B$ be the batch size, in each iteration, batches of $2B$ samples are generated randomly by selecting the same number of source and target samples. The adversarial DA loss is defined using cross-entropy, as follows \cite{ganin2016domain}:
\begin{equation}
\label{eq:total_loss}
\begin{aligned}
   \mathcal{L}_{adv} \, = \, - \frac{1}{2B} \sum_{j=1}^{2B} z_j\log \textbf{D}_i(\bf{I}_j)
   \, + \, (1-z_j)\log \big(1-\textbf{D}_i(\bf{I}_j)\big)
\end{aligned}
\end{equation}
where $z_j = \mathbf{1}(\bf{x}_j \in \mathcal{D}_i)$ the domain label of example $\bf{x}_j$. $\textbf{I}_j$ are the features extracted by $e_f$. The domain classifier $D$ will predict if a given image representation $\bf{I}_j$ is from a \emph{source} ($D_i(\bf{I}_j) =1$) or a \emph{target} domain ($D_i(\bf{I}_j) =0$).

\begin{figure}[h]
    \centering
    \includegraphics[width = 0.47 \textwidth]{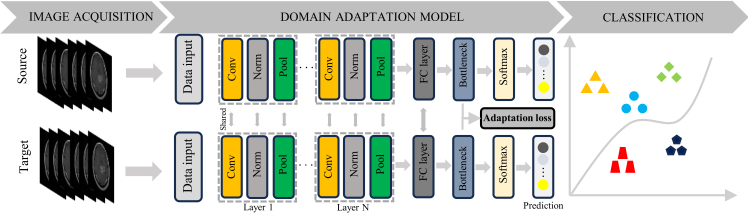}
    \caption{Pipeline of our framework. The model is trained using source data, and the weights of the model are shared to test on the target data. The adaption loss is calculated using the bottleneck layer (256 dimension).}
    \label{fig:PipeLine}
\end{figure}

\mypar{Domain adaptation method for medical purposes}
We demonstrate the application of DA in the medical field using the DANN technique as an example. To align the input size of the neural network, we resize the image to $224 \times 224$. Furthermore, we apply the data normalization method described in \cite{singh2020investigating}. The detailed parameters used in our experiments are provided in Table \ref{T:Details_Alg}. We employ the same approach for the training and testing domains.

\begin{table}[!ht] \scriptsize
 
    \caption{Summary of the implementation details of each algorithm.}
 \setlength{\tabcolsep}{6.5pt}
    \begin{tabular}{ccccccccc}
    \toprule
     Alg. & LR & Epoch & BS &M & OS &WD & Iter \\ 
     \midrule
     DANN  &$1 \times 10^{-2}$ &100 &16&0.6&SGD&$1\times 10^{-3}$&200\\
     w/o DA &$1 \times 10^{-2}$ &100 &16&0.6&SGD&$1\times 10^{-3}$&200\\ 
     \bottomrule
    \end{tabular}\\
    {LR: Learning rate; BS: Batch size; mom: Momentum; OS: Optimization strategy; SGD: Stochastic gradient descent; WD: Weight decay; Iter: iteration per epoch.}
    \label{T:Details_Alg}
\end{table}

\mypar{Domain adaptation training}
During this phase, two major steps are needed: Source domain classification loss backpropagation and DA loss backpropagation. For the classification loss function, we used Cross Entropy (CE) loss, with a SGD optimizer, learning rate = 0.01, weight decay = 0.0005, momentum = 0.9 and batch size = 16 \cite{de2005tutorial,bottou2010large}. For the DA loss, we measure the DANN loss using the features projected by the last fully connected (FC) layer (typically 256 dimension). The whole loss function can be written as:
\begin{equation}
    loss = \mathcal{L}_{CE}(X_s, Y_s) + \mathcal{L}_{adv}(X_s, X_t)
\end{equation}
where the $X_s$, $Y_s$, $X_t$ are the source data, source data's label and target data, respectively.


\begin{table}[ht] \scriptsize
     \setlength{\tabcolsep}{6.9pt}
    \caption{Test accuracy with/without domain adaptation. \textbf{Bold} indicates the best.}
    \begin{tabular}{c|cccc|cccc}
    
    \toprule
     Alg.&SC & BT &MC &CC&SC & BT &MC &CC  \\
     \midrule
    &\multicolumn{4}{c}{\textsc{Alexnet}}&\multicolumn{4}{c}{\textsc{GoogleNet}}\\
    \midrule
\rowcolor{gray!15}    DANN &\textbf{85.4}&\textbf{84.3}&\textbf{86.8}&\textbf{97.9}&86.4&\textbf{87.9}&\textbf{88.3}&97.8  \\
    w/o DA &85.2&79.9&84.8&97.7&\textbf{87.4}&83.8&87.2&\textbf{98.6} \\
    \midrule
    &\multicolumn{4}{c}{\textsc{VGG16}}&\multicolumn{4}{c}{\textsc{ResNet34}}\\
    \midrule 
\rowcolor{gray!15}     DANN&87.4&\textbf{91.5}&\textbf{88.9}&98.2&87.8&\textbf{89.3}&\textbf{89.8}&\textbf{98.6}   \\
    w/o DA&\textbf{88.5}&83.7&87.8&\textbf{98.3}&\textbf{88.7}&84.6&87.7&98.5  \\
    \midrule
    &\multicolumn{4}{c}{\textsc{ResNet50}}&\multicolumn{4}{c}{\textsc{ResNet101}}\\
    \midrule
\rowcolor{gray!15}     DANN&87.5&\textbf{87.9}&\textbf{89.1}&98.2& 87.5&\textbf{89.8}&\textbf{89.2}&97.9   \\
    w/o DA&\textbf{88.3}&84.7&88.1&\textbf{98.4}&\textbf{88.3}&85.1&88.0&\textbf{98.0}  \\
    \midrule
    &\multicolumn{4}{c}{\textsc{DenseNet121}}&\multicolumn{4}{c}{\textsc{ShuffleNet}}\\
    \midrule
\rowcolor{gray!15}     DANN&88.9&\textbf{89.8}&\textbf{90.3}& \textbf{98.9}& 86.7&\textbf{87.0}&\textbf{88.1}&\textbf{97.9}  \\
    w/o DA&\textbf{89.9}&84.9&88.8&98.7&\textbf{87.1}&84.1&86.6&\textbf{97.9}  \\
    \midrule
    &\multicolumn{4}{c}{\textsc{Efficien tNet}}&\multicolumn{4}{c}{\textsc{MnasNet}}\\
    \midrule
\rowcolor{gray!15}     DANN& 86.9&\textbf{89.3}&\textbf{88.5}&\textbf{97.8}& 85.4&\textbf{85.7}&\textbf{87.2}&\textbf{97.4}   \\
    w/o DA&\textbf{87.6}&85.7&87.6&\textbf{97.8}&\textbf{86.1}&83.5&86.6&96.2  \\
    \bottomrule
    \end{tabular}
    {The best result in each dataset is indicated with \textbf{bold} text. SC: Skin cancer; BT: Brain tumor; MC: Multi-cancer; CC: Chest cancer.}
    \label{Performance}
\end{table}

\begin{figure}[!ht]
    \centering
    \includegraphics[width = 0.99 \linewidth]{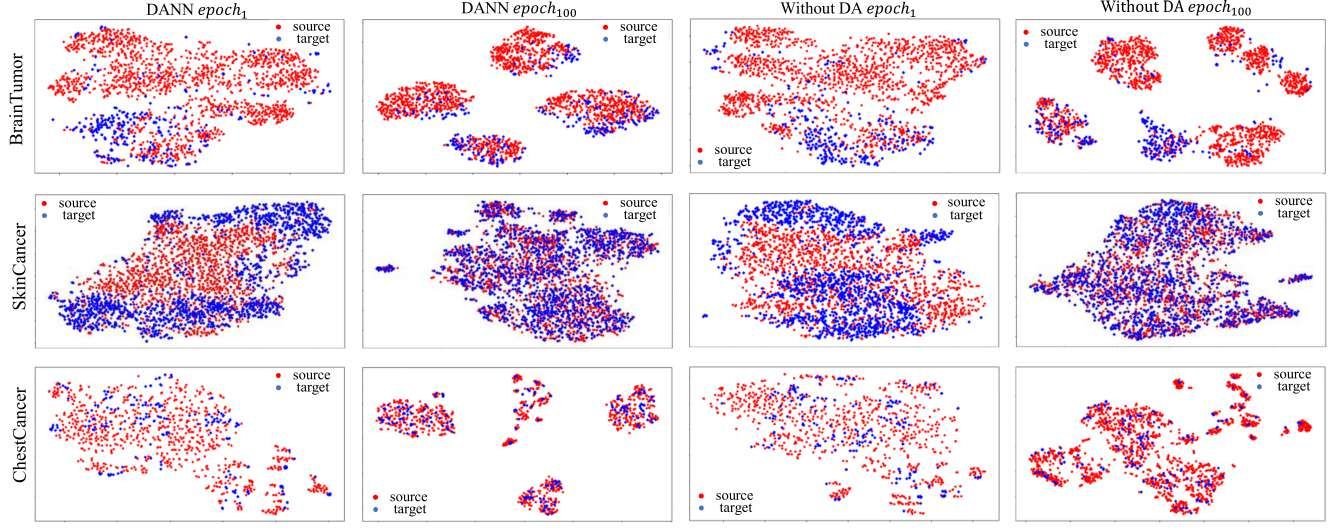}
    \caption{The visualizations of the learned representations using t-SNE for brain tumor, skin cancer and chest cancer. Red points are source
samples and blue are target samples. We use ResNet34 as neural network backbone.}
    \label{fig:DANN_TSNE}
\end{figure}

\section{Experiments}\label{S:3}
We choose classification tasks, the most common application of DA in the medical field, to show the potential of DA. We covered three distinct types of data, including dermoscopic images, magnetic resonance imaging (MRI), and CT images.

\mypar{Datasets} We considered four data sets for evaluation as follows. 1) \textbf{HAM10000 (SC)} dataset \cite{tschandl2018ham10000}, which comprises 10015 dermatoscopic images classified into seven main groups: actinic keratoses and intraepithelial carcinoma / Bowen's disease (akiec), basal cell carcinoma (bcc), benign keratosis-like lesions (bkl), dermatofibroma (df), melanoma (mel), melanocytic nevi (nv), and vascular lesions (vasc). Our training data set consists of 8512 samples, while the test dataset comprises 1503 samples. 2) \textbf{Brain tumor (BT)} dataset \cite{BrainTumor_MRI_1}, which has four different classes, namely glioma tumor, meningioma tumor (megloma), no tumor and pituitary tumor. The training set has 2870 samples, while the test set has 394 samples. 3) \textbf{Chest Xray cancer (CC)} dataset \cite{Chestcancer}, which has four different labels, namely Adenocarcinoma (Adeno, 333 samples), Large cell carcinoma (LCC, 177 samples), Normal (215 samples) and Squamous Cell Carcinoma (SCC, 257 samples). 4) \textbf{Multi-modality dataset (MC).} We combine the previous skin cancer, chest cancer, and brain tumor datasets to test the robustness and generalizability of DA under different image types (e.g., MRI, CT). For all datasets, following \cite{chaddad2023enhancing}, the training set is considered as the source domain, while the test set is used as the target domain.

\begin{figure*}[ht]
    \centering
    \includegraphics[width = 0.97 \linewidth]{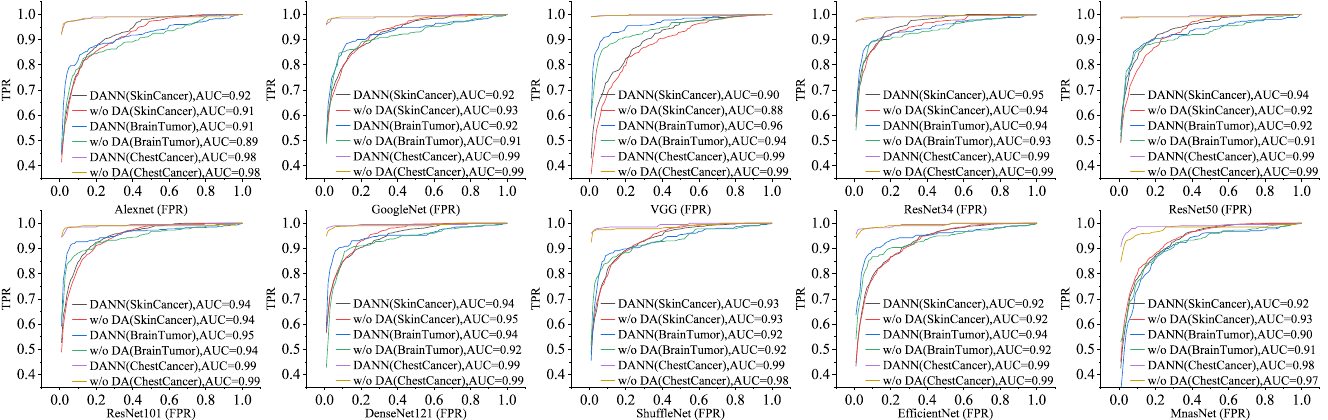}
    \caption{ROC curves with corresponding AUC values in brain tumor, skin cancer and chest cancer datasets. TPR and FPR represent true positive rate and false positive rate, respectively.}
    \label{fig:ROC}
\end{figure*}

\begin{table}[ht] \scriptsize
    \setlength{\tabcolsep}{6.9pt}
    \caption{Global testing accuracy in medical datasets. \textbf{Bold} means the best.}
    \begin{tabular}{c|cccc|cccc}
    
    \toprule
     Alg.&SC & BT &MC &CC&SC & BT &MC &CC  \\
     \midrule
    &\multicolumn{4}{c}{\textsc{Alexnet}}&\multicolumn{4}{c}{\textsc{GoogleNet}}\\
    \midrule
\rowcolor{gray!15}    DANN &{78.4}&\textbf{74.4}&{74.7}&\textbf{96.4}&77.6&{77.4}&\textbf{75.3}&\textbf{93.8}  \\
    w/o DA &\textbf{78.7}&73.8&\textbf{75.0}&\textbf{96.4}&\textbf{78.7}&\textbf{79.9}&73.6&\textbf{93.8} \\
    \midrule
    &\multicolumn{4}{c}{\textsc{VGG16}}&\multicolumn{4}{c}{\textsc{ResNet34}}\\
    \midrule 
\rowcolor{gray!15}     DANN&79.5&{77.2}&{76.5}&\textbf{98.5}&81.0&{81.5}&\textbf{77.8}&\textbf{97.4}   \\
    w/o DA&\textbf{80.0}&\textbf{80.4}&\textbf{76.7}&{98.4}&\textbf{81.6}&\textbf{81.9}&77.6&\textbf{97.4}  \\
    \midrule
    &\multicolumn{4}{c}{\textsc{ResNet50}}&\multicolumn{4}{c}{\textsc{ResNet101}}\\
    \midrule
\rowcolor{gray!15}     DANN&\textbf{82.5}&\textbf{82.5}&\textbf{78.7}&\textbf{95.4}& 82.6&{82.0}&\textbf{79.7}&\textbf{97.9}   \\
    w/o DA&{81.2}&81.2&\textbf{78.7}&{95.3}&\textbf{83.7}&\textbf{82.2}&79.4&\textbf{97.9}  \\
    \midrule
    &\multicolumn{4}{c}{\textsc{DenseNet121}}&\multicolumn{4}{c}{\textsc{ShuffleNet}}\\
    \midrule
\rowcolor{gray!15}     DANN&\textbf{83.3}&{79.2}&\textbf{80.2}& \textbf{97.9}& \textbf{79.3}&\textbf{77.4}&{77.4}&\textbf{96.4}  \\
    w/o DA&{83.0}&\textbf{81.7}&79.0&\textbf{97.9}&{78.9}&77.1&\textbf{77.8}&\textbf{96.4}  \\
    \midrule
    &\multicolumn{4}{c}{\textsc{EfficientNet}}&\multicolumn{4}{c}{\textsc{MnasNet}}\\
    \midrule
\rowcolor{gray!15}     DANN& \textbf{76.0}&{76.4}&\textbf{74.3}&\textbf{95.9}& \textbf{69.2}&{48.2}&\textbf{63.6}&\textbf{67.7}   \\
    w/o DA&\textbf{76.0}&\textbf{78.4}&73.3&\textbf{95.9}&\textbf{69.2}&\textbf{48.7}&63.3&67.6  \\
    \bottomrule
    \end{tabular}
    {The best result using different CNNs is indicated with \textbf{bold} text. SC: Skin cancer; BT: Brain tumor; MC: Multi-cancer; CC: Chest cancer.}
    \label{Performance:FL}
\end{table}

\mypar{Experimental settings}
To provide a practical simulation of DA techniques in the medical field, we have selected popular neural network models of the last decade as backbone for testing. The neural networks we selected are Alexnet \cite{kennedy2012candishare}, GoogleNet \cite{szegedy2015going}, VGG16 \cite{simonyan2014very}, Resnet34, Resnet50, Resnet101 \cite{he2016deep}, Densenet121 \cite{huang2017densely}, ShuffleNet \cite{zhang2018shufflenet}, EfficientNet-b2 \cite{tan2019efficientnet}, and MnasNet \cite{tan2019mnasnet}. The training epoch is set to 100. Figure \ref{fig:PipeLine} shows a pipeline of our proposed DA framework. We calculate the loss of adaptation in the bottleneck layer. The testing environment used for this study is based on the Windows 11 operating system, featuring an Intel 13900KF CPU, 128 GB of RAM, and an RTX 4090 graphic card. We use PyTorch 1.12.0 with Python 3.7. We measure the classification ability of the models using the accuracy (ACC) and receiver operating characteristic (ROC) values with the area under the ROC curve (AUC). Furthermore, the expected calibration error (ECE) and reliability diagram \cite{10902405} are used to evaluate the calibration effects. ECE quantifies calibration accuracy by showing how closely predicted probabilities match true outcomes. Reliability plots visually represent this alignment, providing intuitive insight into model confidence. Since the target domain data are fixed, we set the same random seed (42) for all networks to ensure a fair comparison.

\subsection{Experimental results} 
Table \ref{Performance} reports the testing accuracy using DA and without DA for these datasets. We observe that 1) DA for skin cancer indicates nearly identical accuracy with CNN alone (e.g., 85.4\%  with DA vs 85.2\%  without DA using AlexNet). This suggests a negative transfer process, which implies that dermatology-related studies may require the design of more appropriate DA methods. 2) For MRI images in BT, the use of DA leads to significant accuracy boost (e.g., $\sim 5\%$ improvement using AlexNet), especially for VGG16, ResNet34, indicating the need to use DA for the MRI data. 3) For small data sets such as Chest X-ray dataset, the potential of DA is limited, both models with and without DA achieved similar results for all neural networks. and 4) For multi-modal dataset, the models exhibited superior accuracy compared to the models that used CNN only on all networks (e.g., 89.8\%  with DA vs 87.7\%  without DA). We suggest that DA has the ability to better calibrate the network to properly identify different cancer classes.

\mypar{ROC curves and AUC values} Furthermore, we performed ROC analysis with corresponding AUC values for DANN and without DA in brain tumor, skin cancer, and chest cancer datasets. To simplify, we use one-vs-all method and using the unweighted mean value to measure the ROC curves and AUC values. Figure \ref{fig:ROC} shows the ROC curves for CNN models. As illustrated, the use of DA leads to higher AUC values compared to CNN alone in skin cancer for deep models such as AlexNet and ResNet34. Furthermore, the use of DA improves the AUC value for BT with all deep models, which is consistent with the highest testing accuracy as reported in Table \ref{Performance}.

\begin{figure}
    \centering
    \includegraphics[width=0.97\linewidth]{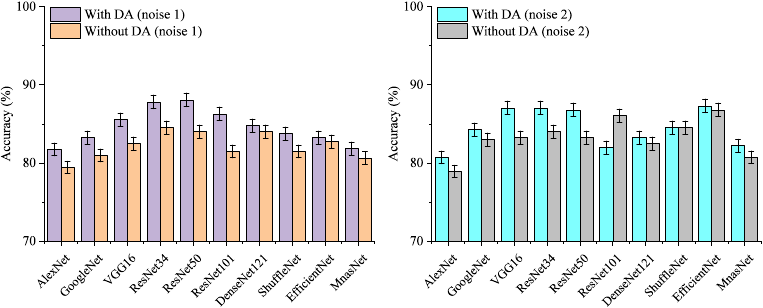}
    \caption{Test accuracy (AVG $\pm$ STD) of 10 deep models on BT dataset (noisy training set).}
    \label{fig:NoisyData}
\end{figure}

\mypar{The impact of noise} As the quality of the source data can affect the adaptation effects of DA \cite{han2021towards}, we further validate the usefulness of DA when training data are noisy. The BT dataset is used for the evaluation. To simulate noisy training data, we designed two noises: 1) Gaussian blur only (kernel size = 3), and 2) Gaussian noise $\mathcal{N}\sim(0,1)$, salt-pepper noise ($p=0.02$), and Poisson noise together. Since the noise is added randomly (i.e., controlled by random seeds), we repeat the experiments three times and report the average (AVG) $\pm$ standard deviation (STD). Figure \ref{fig:NoisyData} illustrates the test accuracy (AVG $\pm$ STD) of 10 deep models using DA and without DA. The results demonstrate that DA effectively alleviates the impact of various types of noisy data, leading to a considerable improvement in model performance (e.g., an increase of $\sim 3\%$ for VGG16). However, for certain models such as ResNet101, the use of DA leads to negative adaptation (e.g., $\sim 5\%$ accuracy drop using noise 2). This suggests that inappropriate network architectures can decrease the potential of DA module.

\begin{figure}[!ht]
    \centering
    \includegraphics[width=0.225\linewidth]{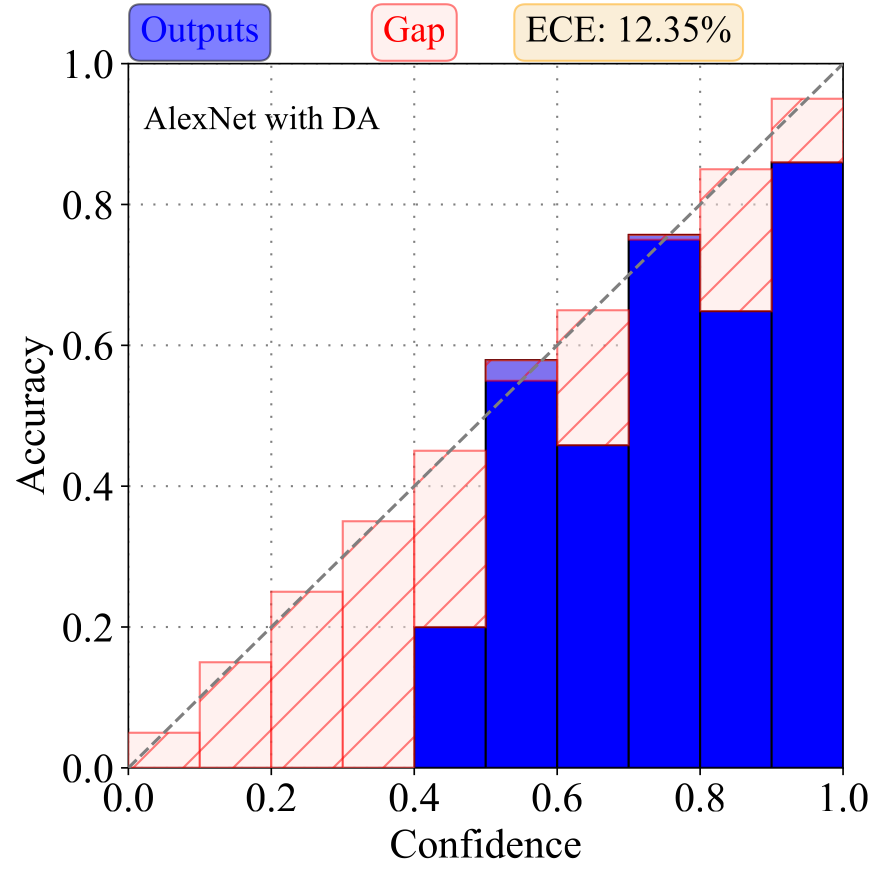} \includegraphics[width=0.225\linewidth]{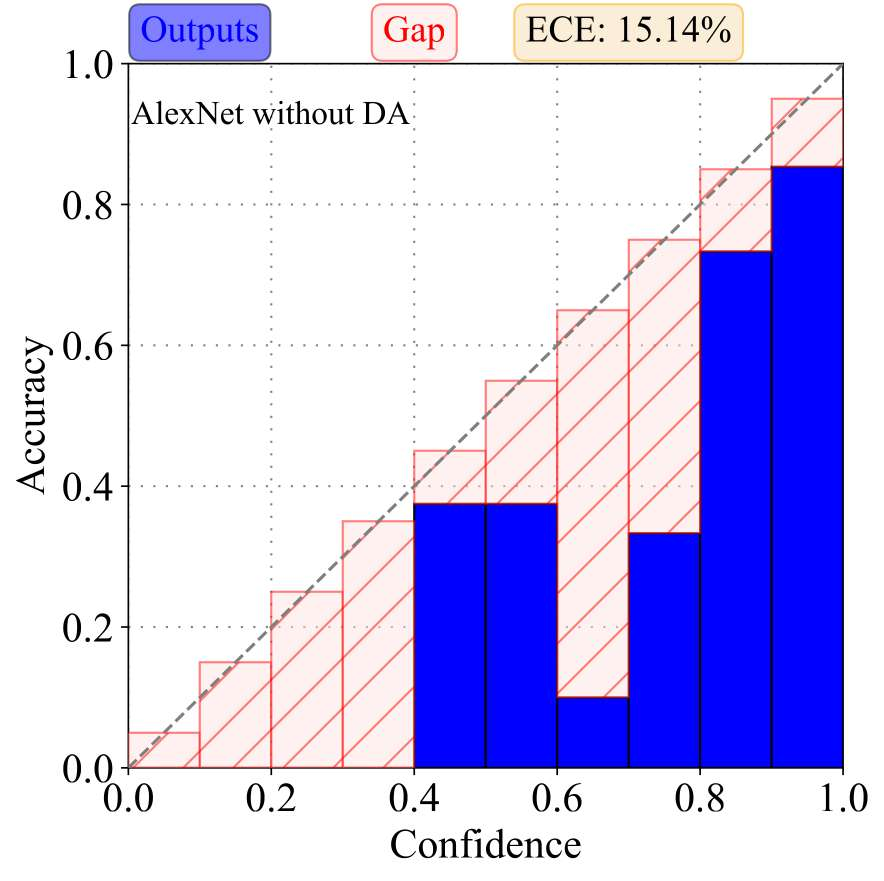} \includegraphics[width=0.225\linewidth]{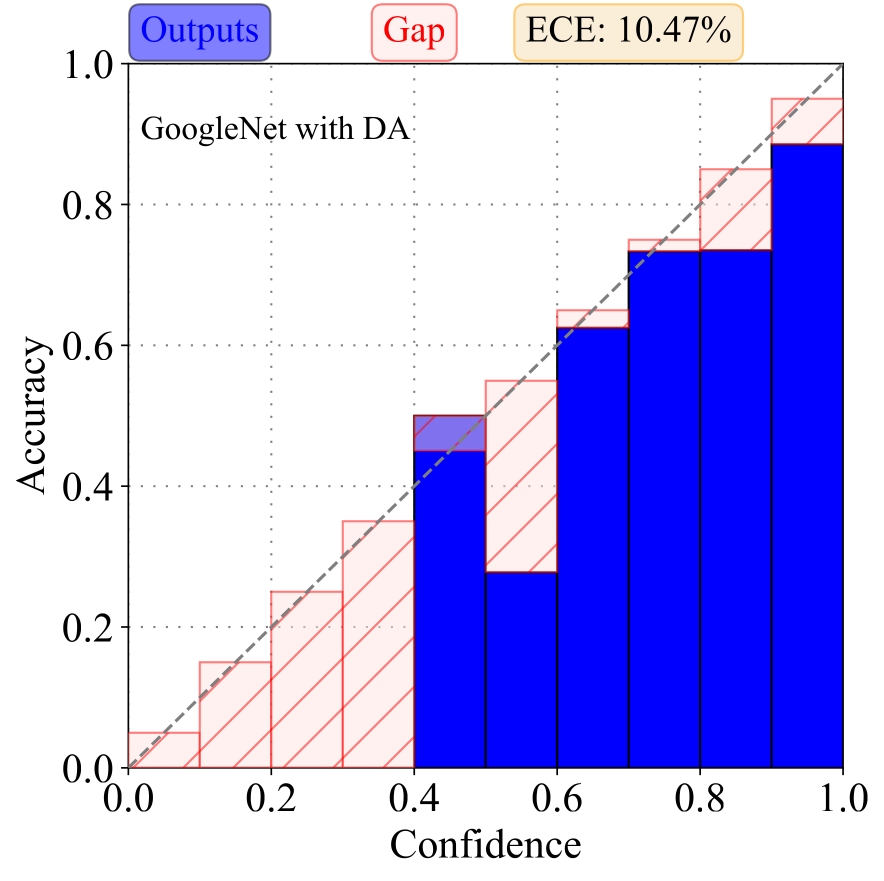} \includegraphics[width=0.225\linewidth]{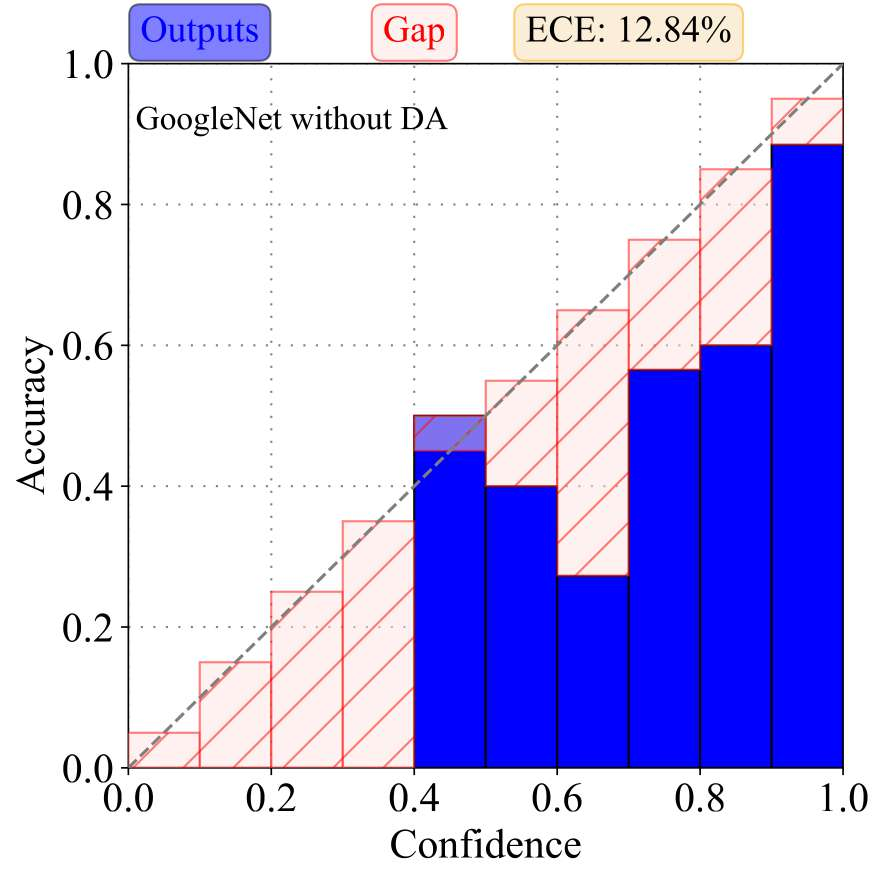}\\
    \includegraphics[width=0.225\linewidth]{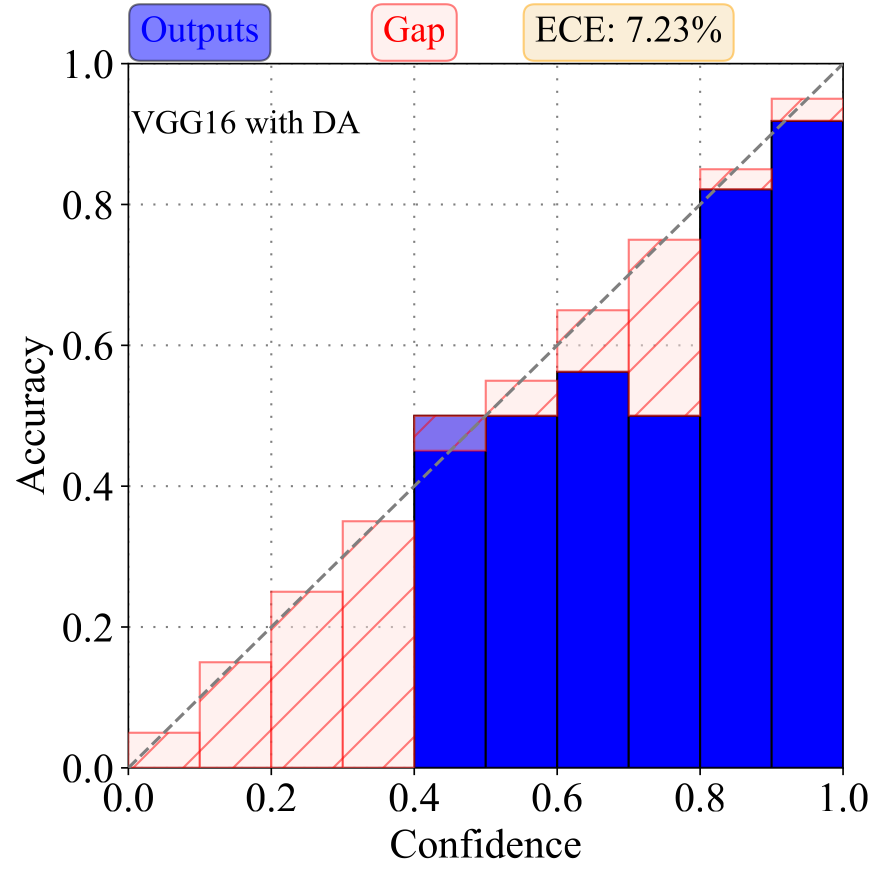} \includegraphics[width=0.225\linewidth]{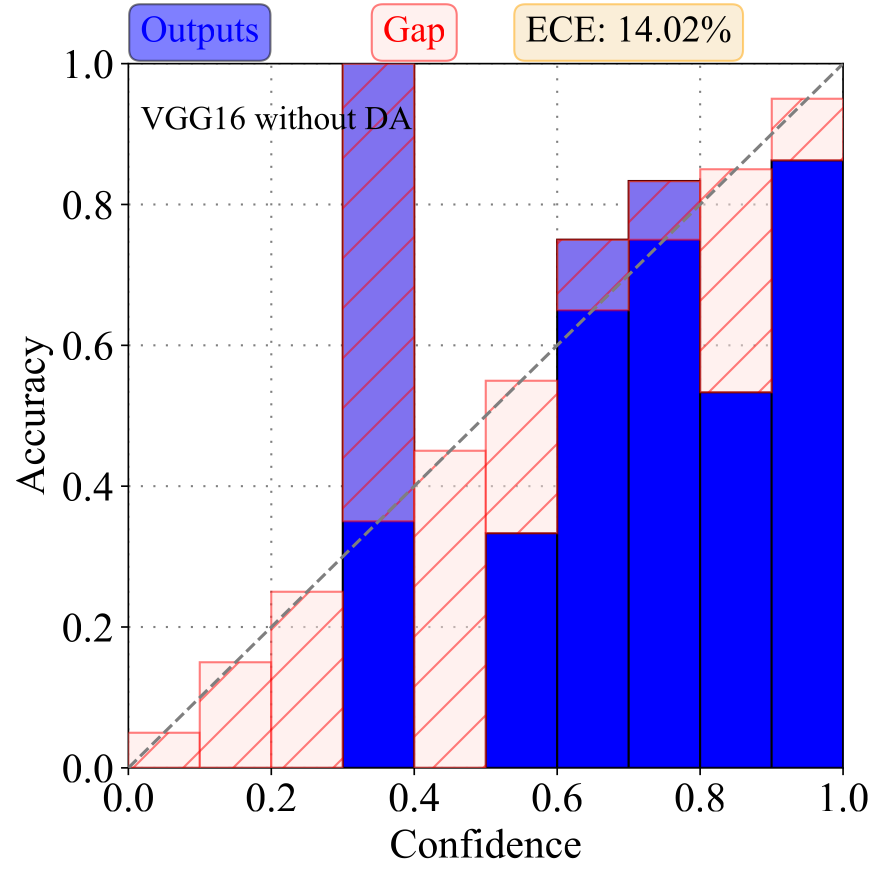}\includegraphics[width=0.225\linewidth]{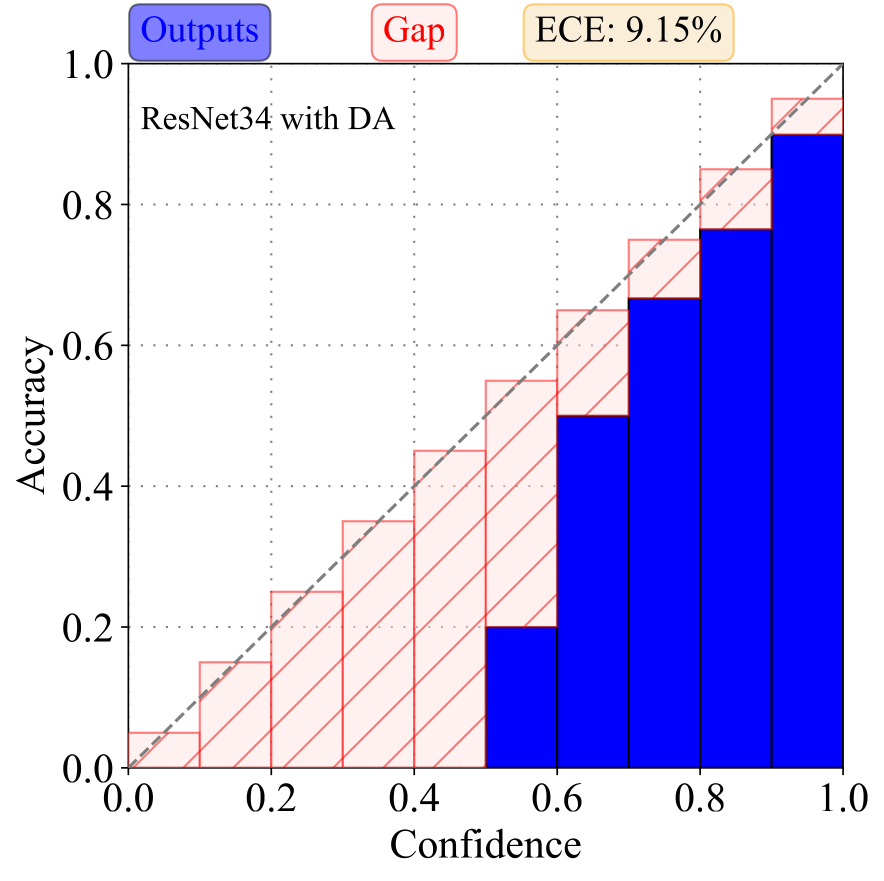} \includegraphics[width=0.225\linewidth]{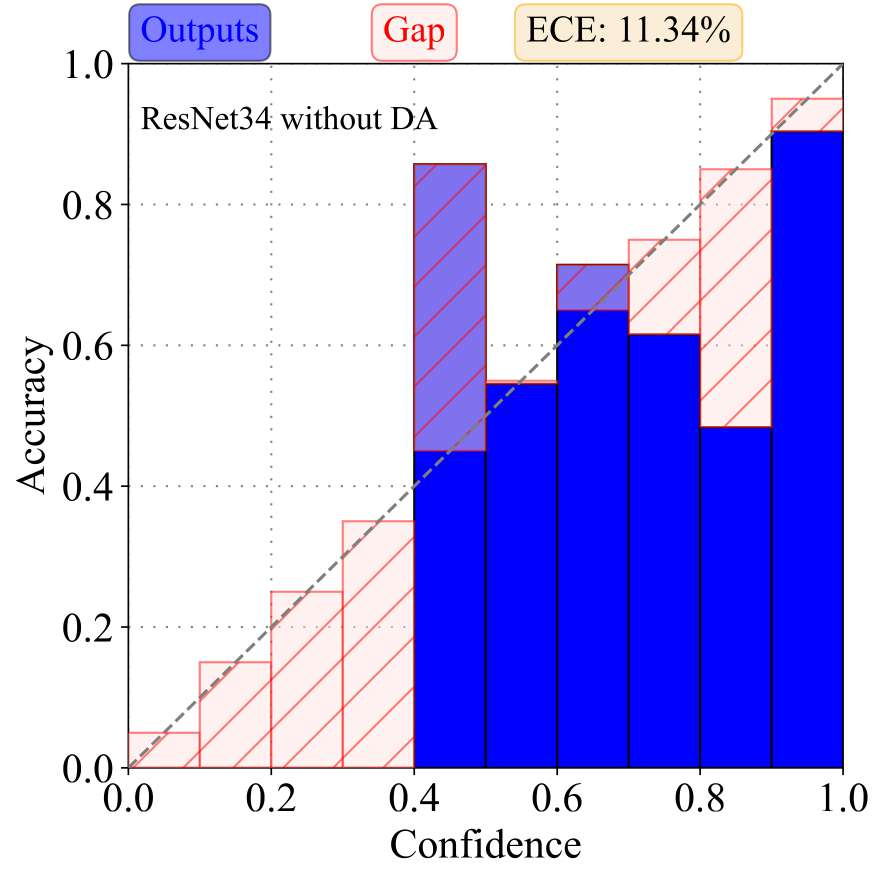}\\
    \includegraphics[width=0.225\linewidth]{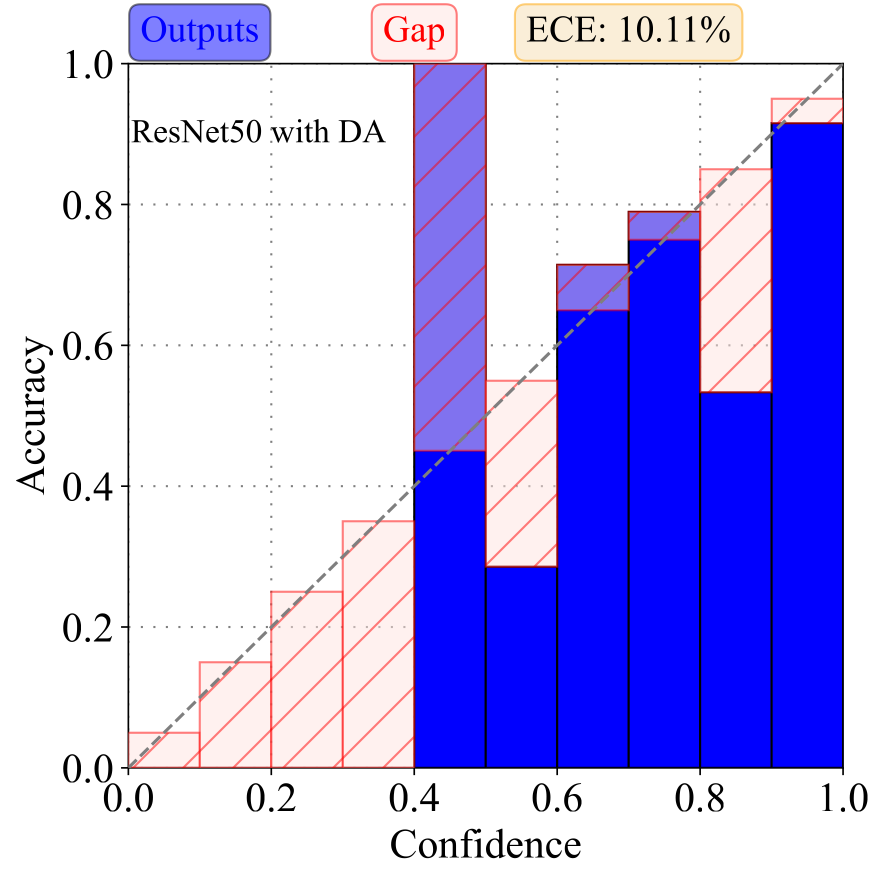} \includegraphics[width=0.225\linewidth]{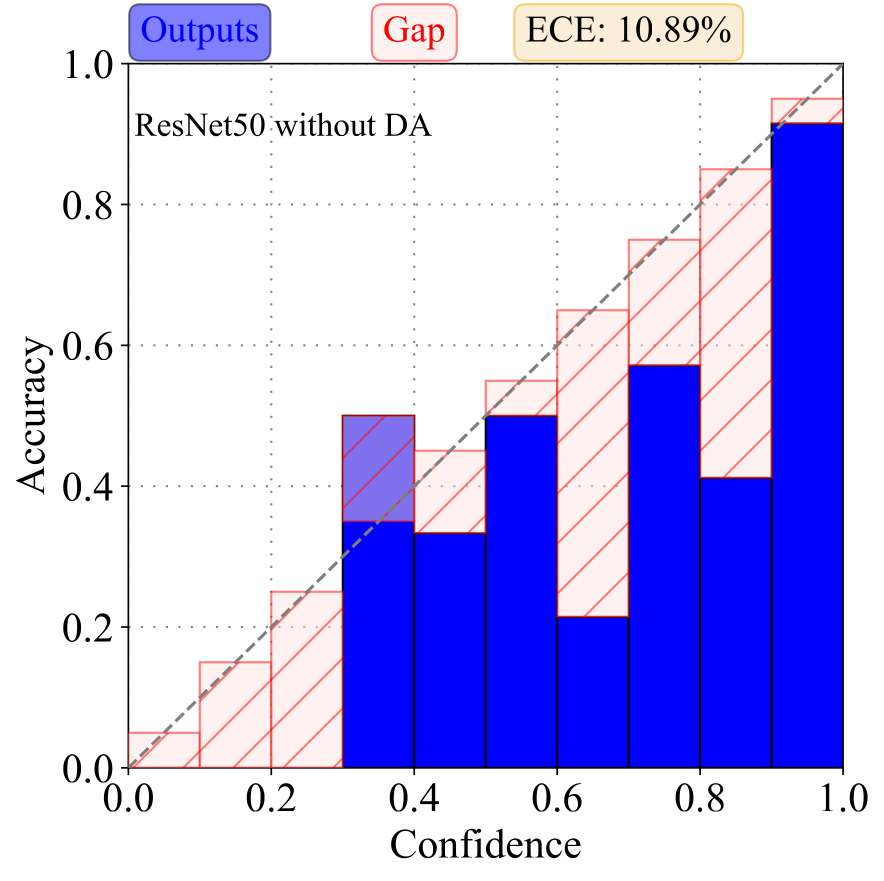}\includegraphics[width=0.225\linewidth]{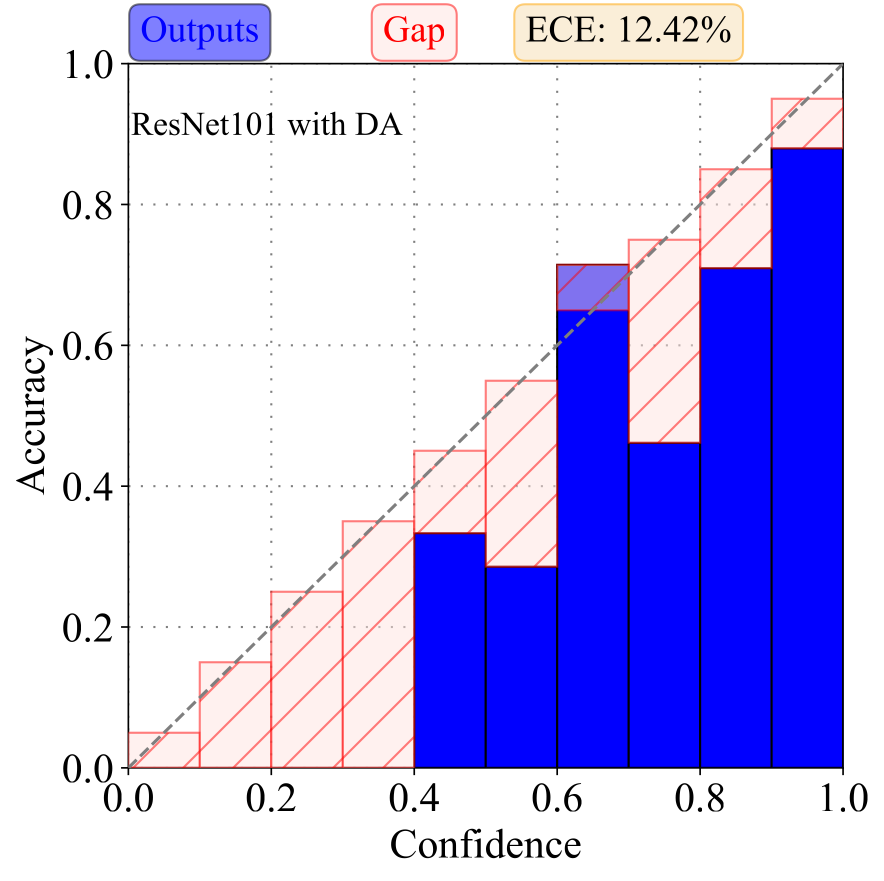} \includegraphics[width=0.225\linewidth]{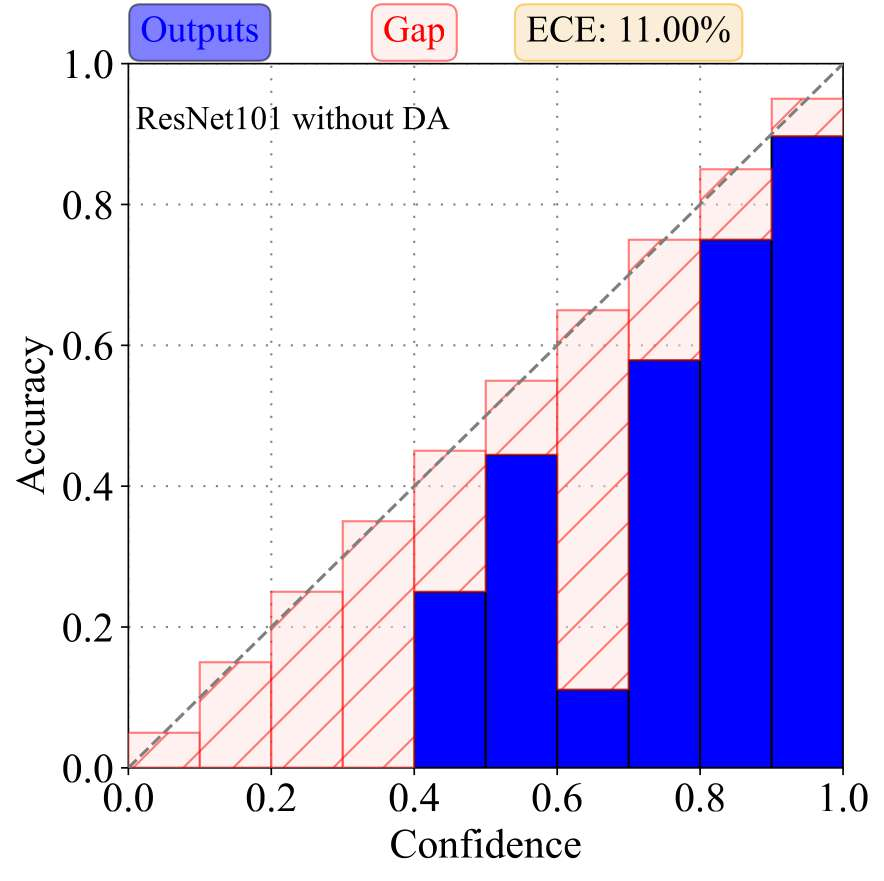}\\
    \includegraphics[width=0.225\linewidth]{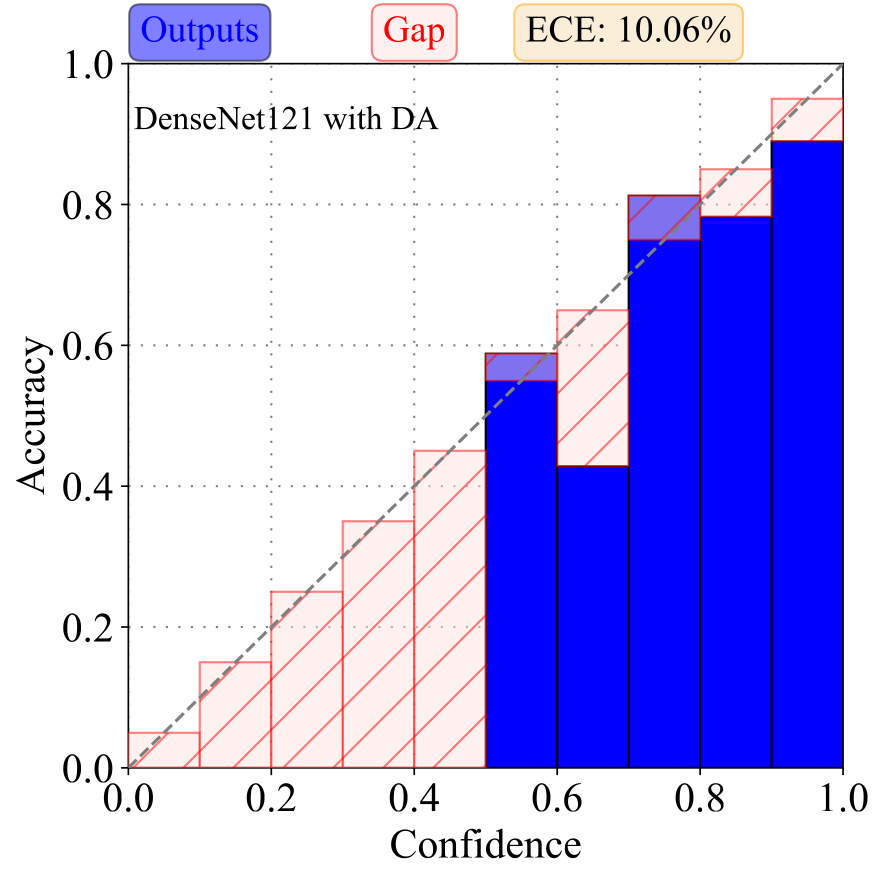} \includegraphics[width=0.225\linewidth]{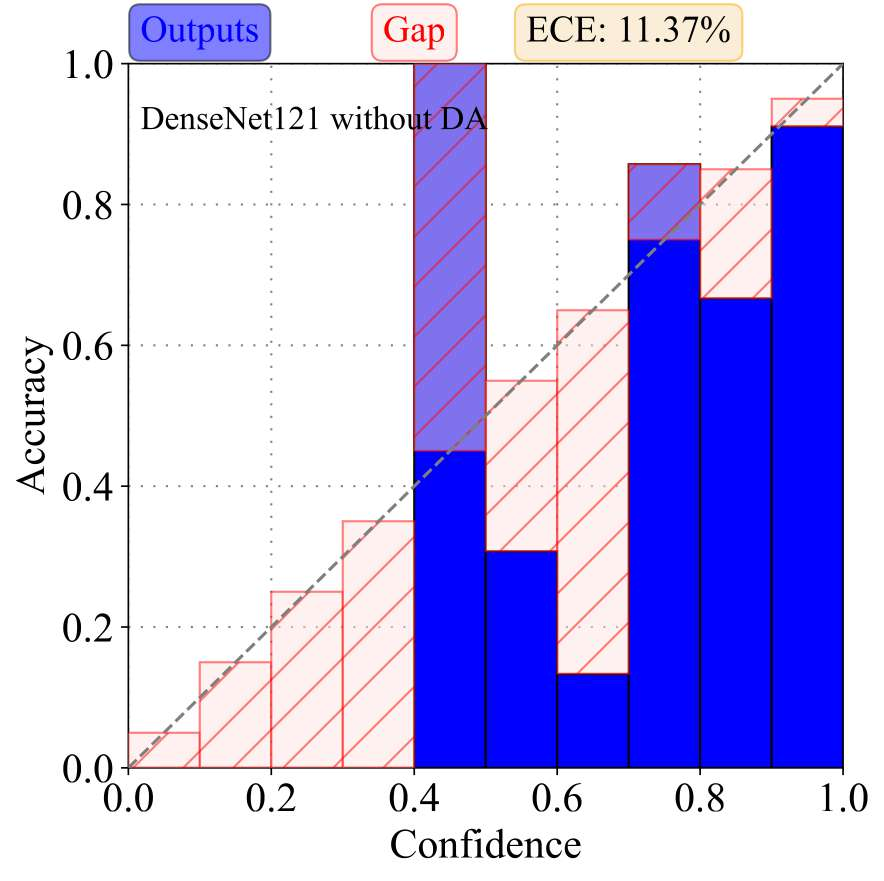}\includegraphics[width=0.225\linewidth]{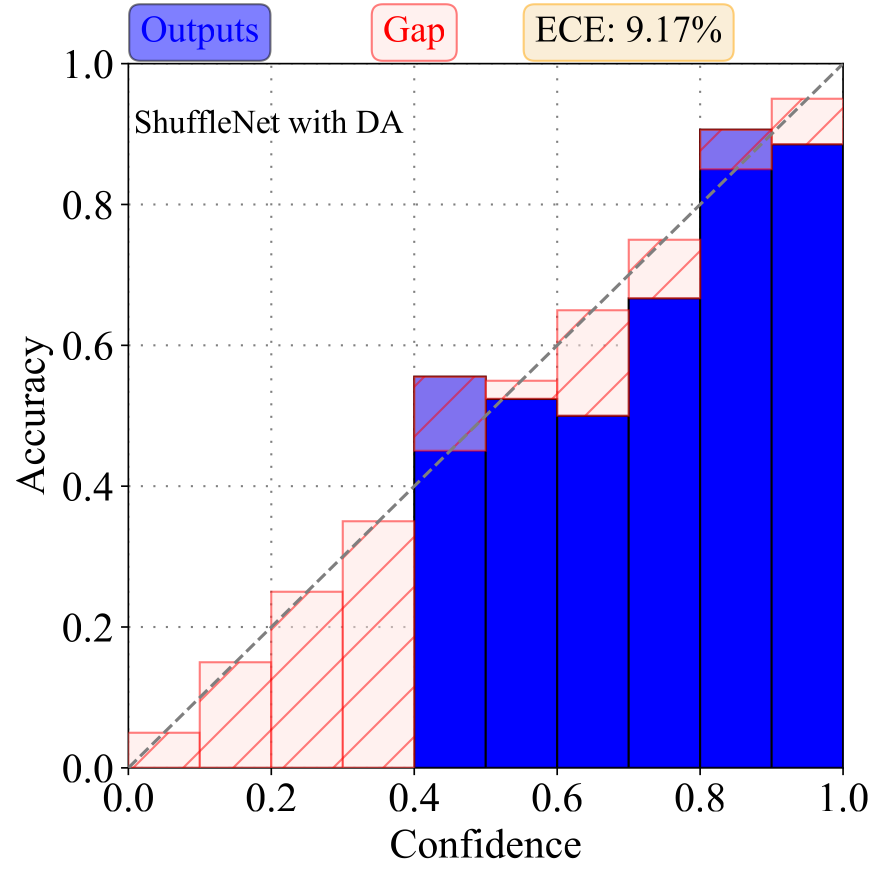} \includegraphics[width=0.225\linewidth]{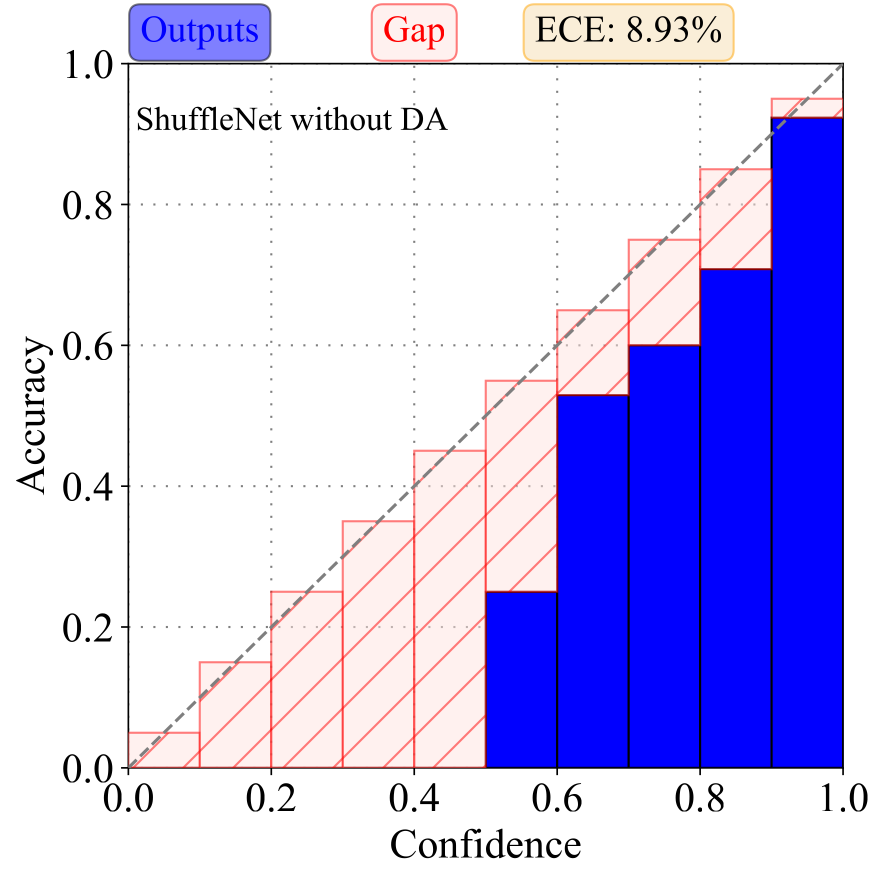}\\
    \includegraphics[width=0.225\linewidth]{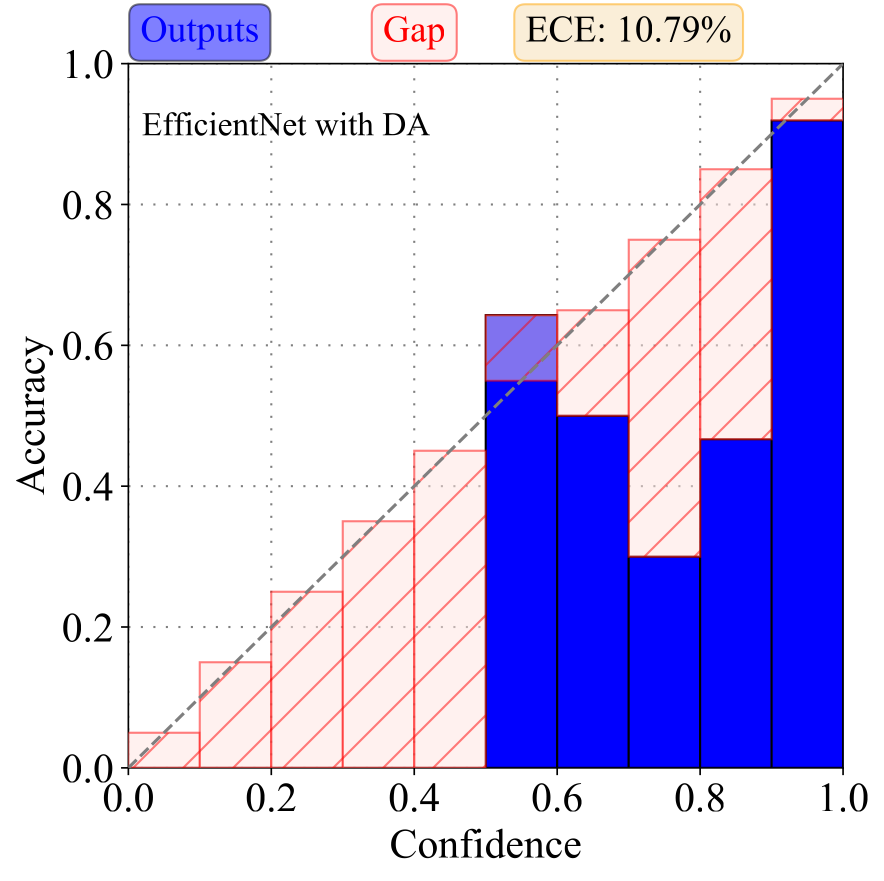} \includegraphics[width=0.225\linewidth]{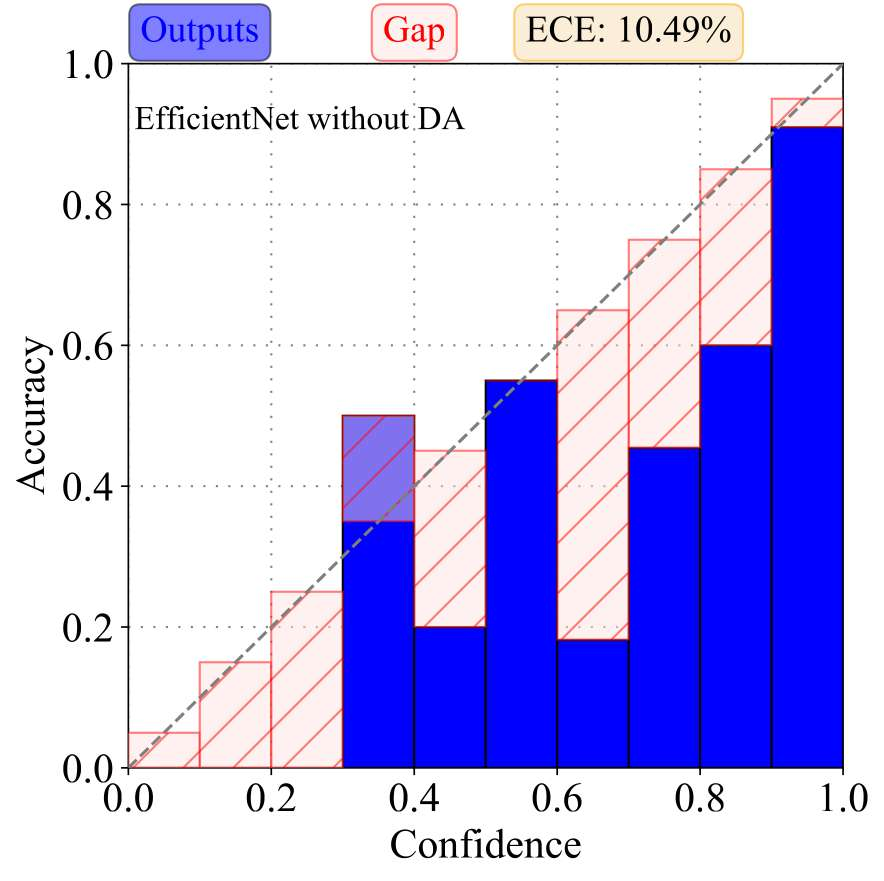}\includegraphics[width=0.225\linewidth]{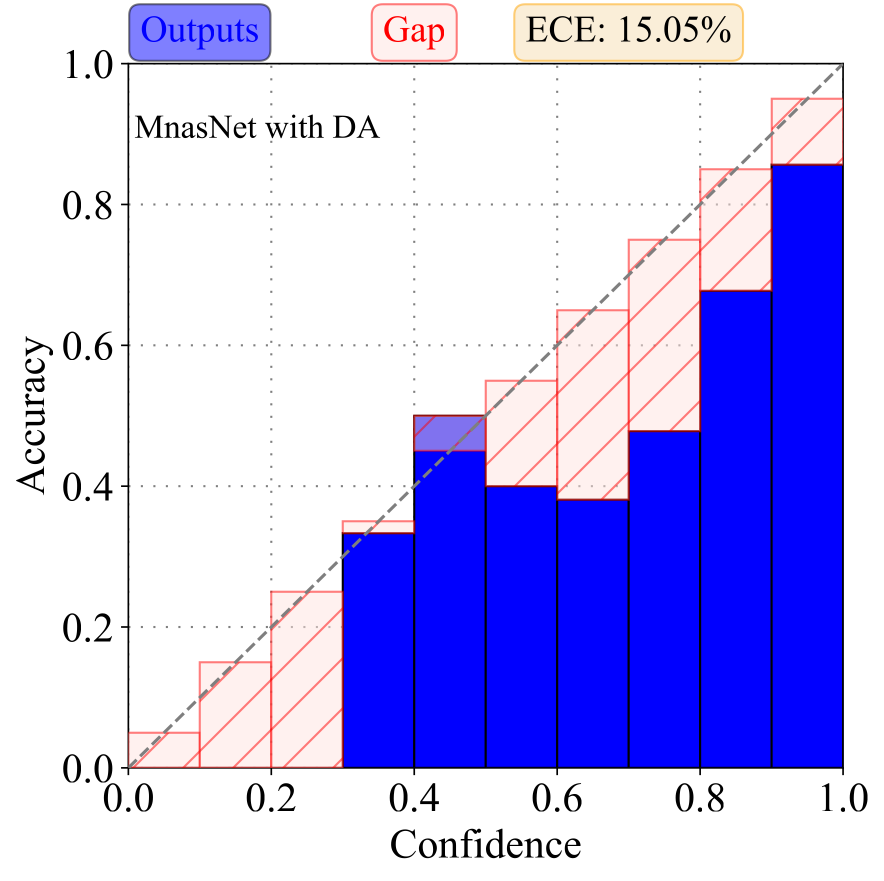} \includegraphics[width=0.225\linewidth]{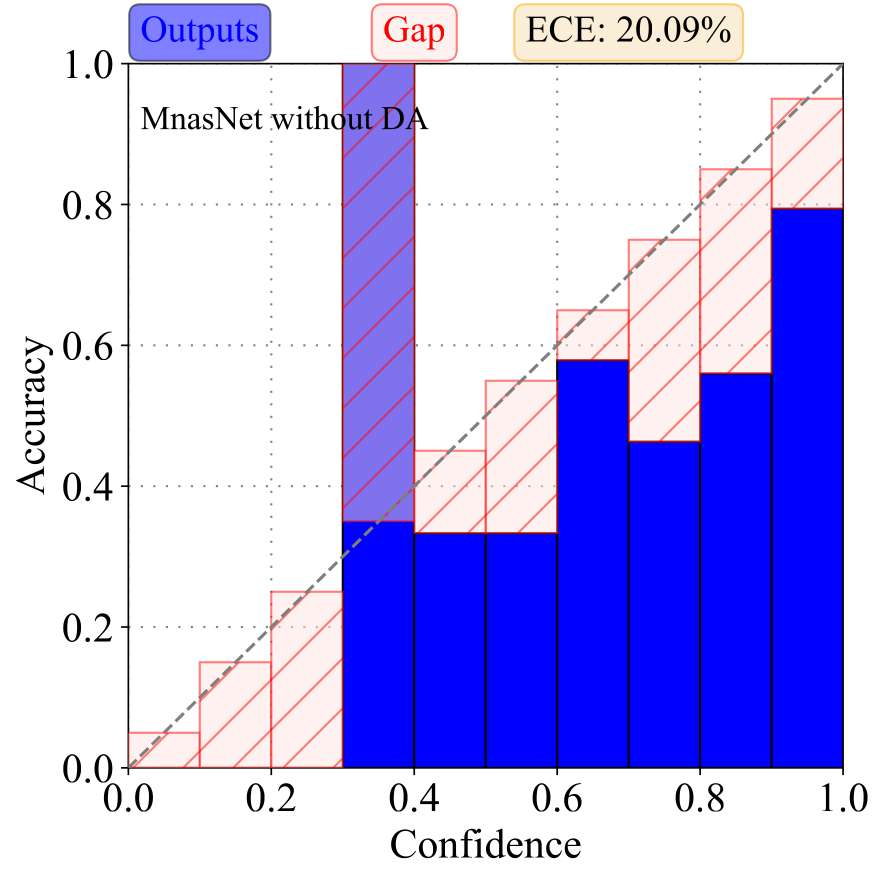}\\
    
    \caption{Reliability diagram with ECE value of 10 deep models using DA and without DA on Brain Tumor dataset.}
    \label{fig:Calibration_BrainTumor}
\end{figure}

\begin{figure}[!ht]
    \centering
    \includegraphics[width=0.8\linewidth]{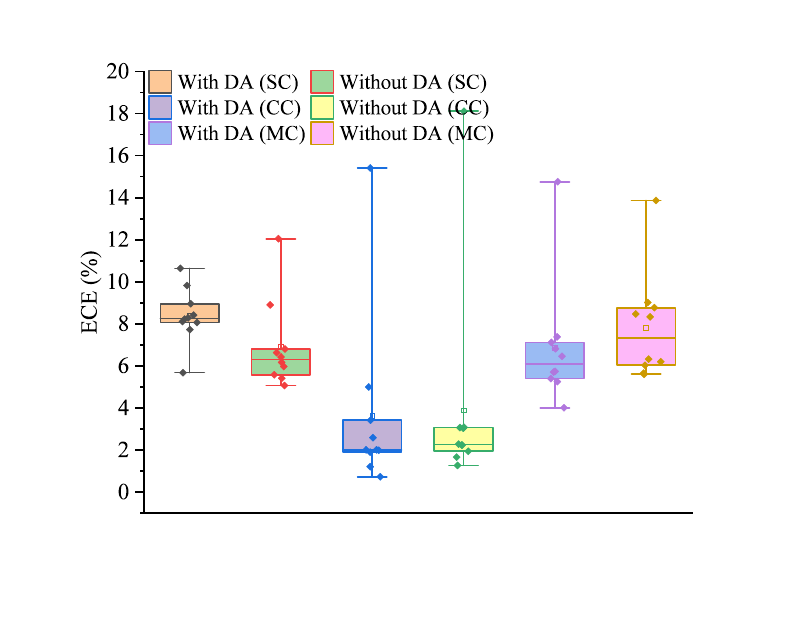}
        \caption{ECE value of 10 deep models using DA and without DA on SC, CC and MC datasets.}
    \label{fig:Calibration_SkinCancer}
\end{figure}

\subsection{t-SNE visualization}  
We also examined the feature adaptation ability for the DA technique using the t-Distributed Stochastic Neighbor Embedding (t-SNE) method \cite{van2008visualizing}. t-SNE is a dimensionality reduction technique used to visualize high-dimensional data by projecting them into a lower-dimensional space (typically 2D) while preserving the local relationships and structure between data points and is widely used to visually demonstrate the effects of DA. We highlight that the DA method has the ability to reduce intra-class discrepancies and maximize extra-class distances, resulting in a distinct boundary for each subclass, as illustrated in Figure \ref{fig:DANN_TSNE}. For example, introducing DA for BT dataset leads to four distinct clusters where the source and target data are well calibrated (DANN $epoch_{100}$), which is consistent with four different classes. However, without using DA, it shows more than four clusters, which is hard to interpret (Without DA $epoch_{100}$). Furthermore, the use of DA in the chest X-ray dataset can provide feasible feature alignments, while using CNN alone indicates poor class calibration ability. For SC dataset, introducing DA leads to seven clusters, while the original CNN shows nearly three clusters. These findings suggest that DA substantially improves feature adaptation for brain and skin cancer datasets, while its minimal impact on chest x-rays reveals limitations in its generalizability in medical imaging domains.


\begin{figure}[!ht]
    \centering
    \includegraphics[width = 0.95 \linewidth]{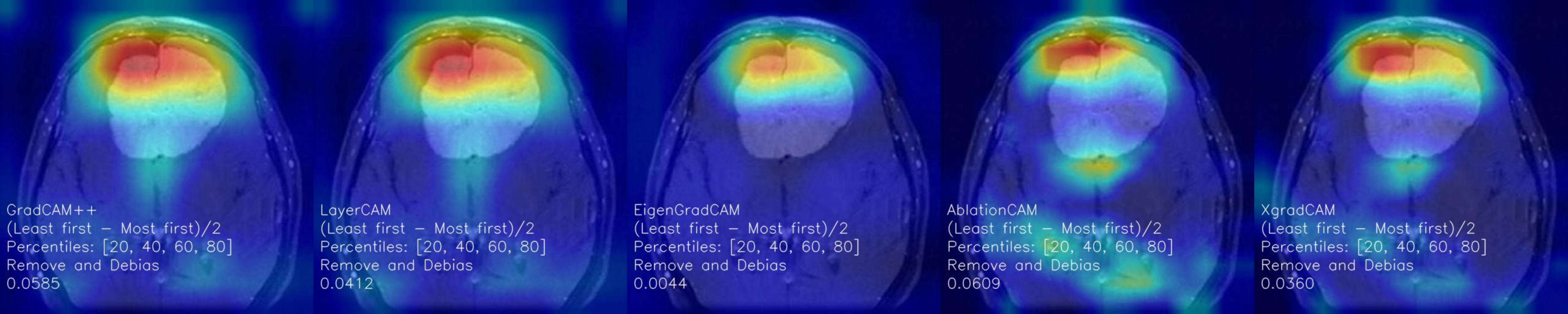}\\
    \vspace{2pt}
    \includegraphics[width = 0.95 \linewidth]{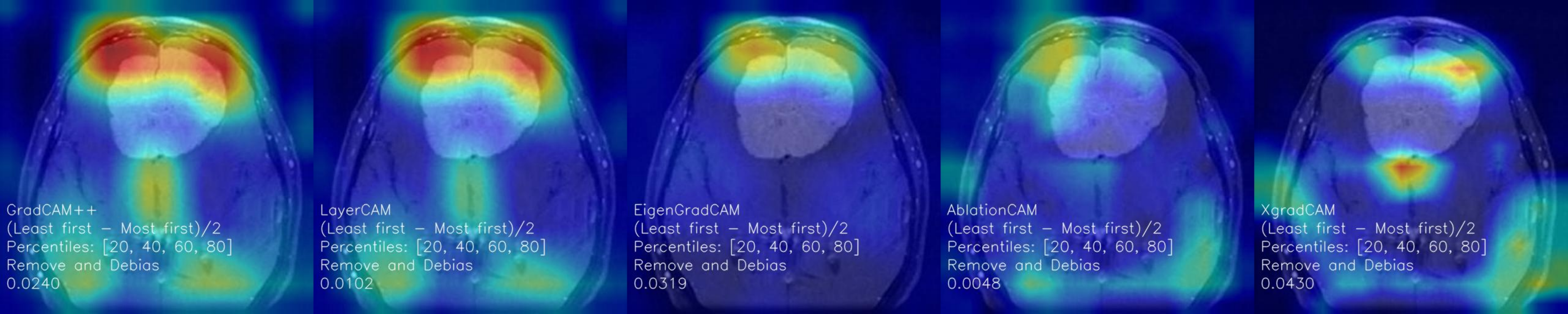}\\
    \vspace{1pt}
    \includegraphics[width = 0.95 \linewidth]{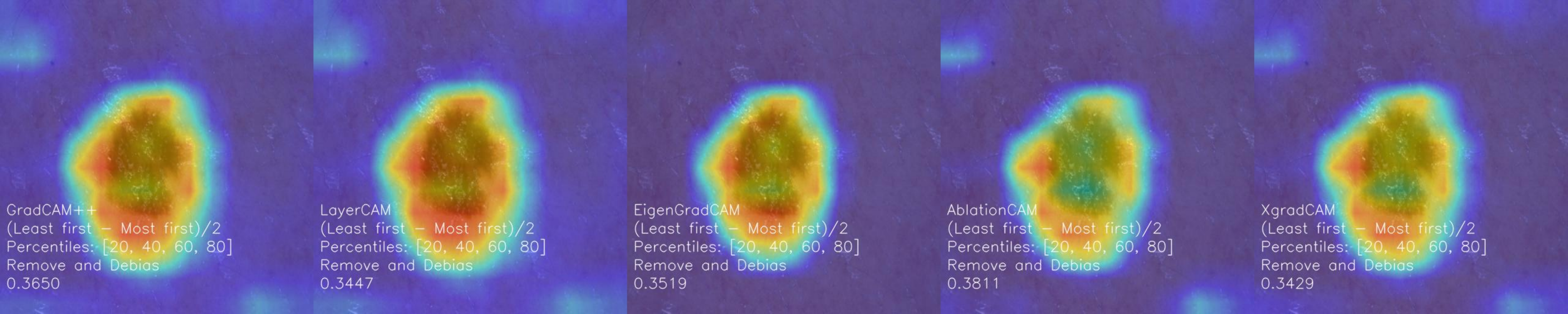}\\
    \vspace{2pt}
    \includegraphics[width = 0.95 \linewidth]{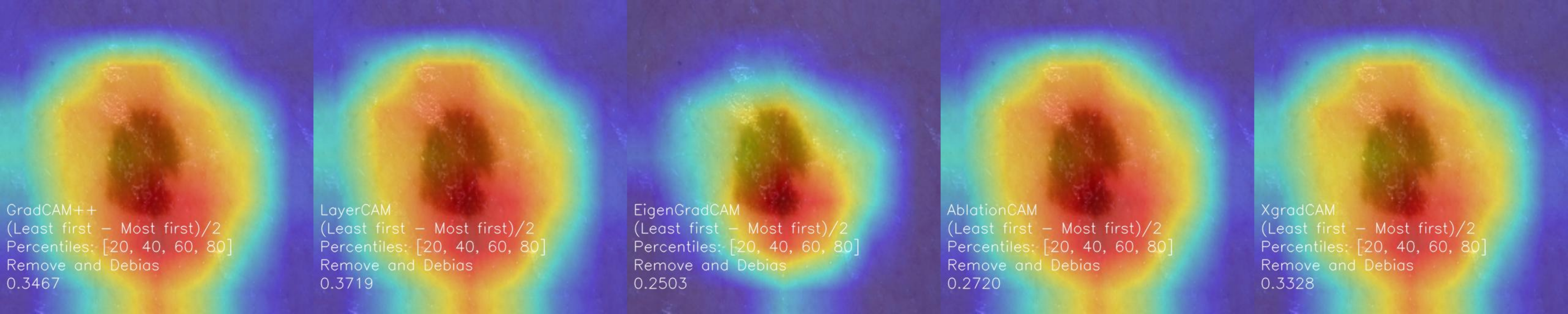}
    \vspace{1pt}
    \includegraphics[width = 0.95 \linewidth]{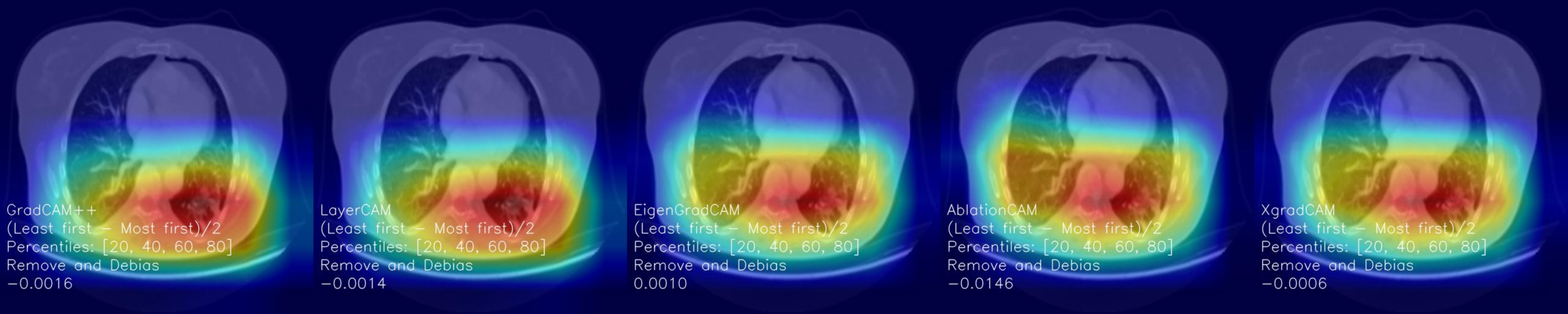}\\
    \vspace{2pt}
    \includegraphics[width = 0.95 \linewidth]{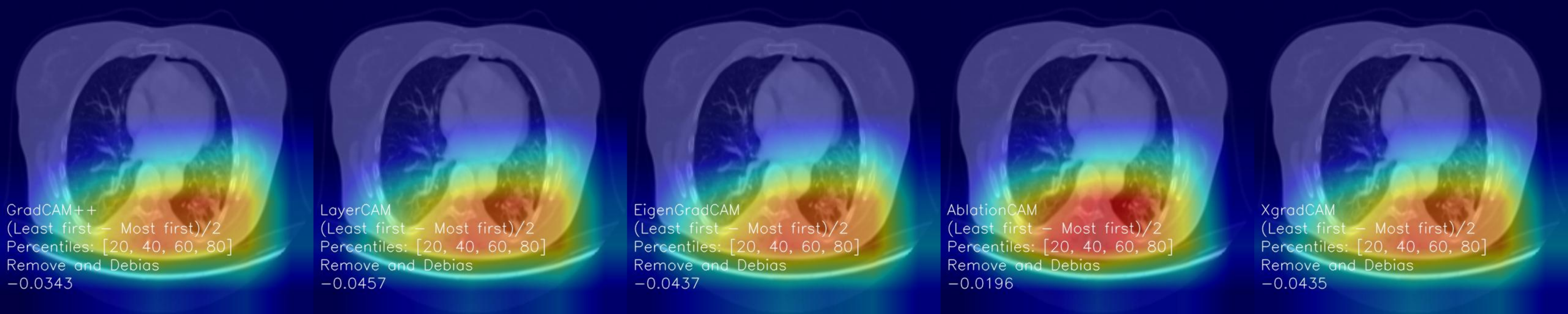}
    \caption{The heatmap visualization with confidence change for DA (\{1,3,5\} row) and without DA (\{2,4,6\} row) on BT, SC and CC datasets. The brightness of the color correlates with the level of the model's attention, with a deeper red indicating a higher level of focus and a lighter blue showing a lack of attention.}
    \label{fig:XAI}
\end{figure}

\subsection{Federated learning} 
We investigate the impact of DA with FL. Following \cite{10288131}, we split the training set in each dataset into three clients with randomly selected samples, while the original testing set is used for global evaluation. We used federated averaging aggregation technique as described in  \cite{10288131} to obtain the global model. The total communication round was set to 50. Table \ref{Performance:FL} reports the global testing accuracy using DANN with 10 CNNs. Using FL indicates accuracy drop compared to centralized approaches as reported in Table \ref{Performance}. Furthermore, using DA in FL can lead to improve the accuracy using ResNet50 in BT dataset. However, for AlexNet, GoogleNet and DenseNet121, the DANN is not effective compared to CNN alone. This suggests that the limited samples in clients can not lead to adaptation effects, thereby decreases the performance of deep models. Furthermore, for large-scale datasets such as MC, using DA can improve deep model accuracy (e.g., GoogleNet with 75.3\%), highlighting the potential of DA in FL.

\subsection{Interpretability analysis} We selected ResNet34 as the network backbone, using skin cancer, brain tumor, chest cancer and multi-modality dataset, to investigate whether DA may improve the interpretability of the model while maintaining high classification accuracies. We use Gradcam++, LayerCAM, EigenGradCAM, AblationCAM and XgradCAM as the XAI methods. Furthermore, to provide a comprehensive evaluation as suggested in \cite{chaddad2024generalizable}, we introduced a quantitative metric named remove and debias (ROAD) \cite{rong2022consistent} to validate the DA model. The main idea of ROAD is to remove part of the image and then predict it again to measure the confidence change (higher is better).

Figure \ref{fig:XAI} illustrates the heatmap visualization obtained using five XAI methods and the confidence change (shown in the image as a number below the Remove and Debias symbols.) using ROAD for DA and without DA models on BT, SC and CC. We observe that: 1) for brain tumor images, the use of DA leads to better visual results as the model focuses more on the abnormal regions; In addition, the confidence change of DA model is higher compared to without DA (e.g., 0.0585 vs 0.0240 using GradCAM++). 2) for skin cancer images, the use of DA let the model focus more on the abnormal regions, while all XAI methods indicate higher overall confidence change. 3) for chest cancer, the potential of DA is limited, as it provides similar visual explanations and confidence changes compared to without DA (e.g., -0.0146 vs -0.0196). These findings suggest that DA can enhance diagnostic reliability for BT and SC by improving the focus of the model on abnormal regions and increasing confidence in predictions, but its limited impact on chest cancer highlights the need for domain-specific adaptations to ensure clinically significant improvements in all medical imaging tasks.



\subsection{Calibration analysis}
We select SC, BT, CC and MC datasets using 10 deep models with and without DA to evaluate their calibration ability. Figure \ref{fig:Calibration_BrainTumor} illustrates the reliability plot with ECE value for 10 deep models on Brain Tumor dataset. Despite the use of ResNet101, ShuffleNet and EfficientNet lead to higher ECE value ranging from [0. 24\% to 1. 42\%], the rest of the deep models indicate better calibration results compared to without DA (e.g., $\sim 7\%$ decrease in ECE value using VGG16). Furthermore, as shown in Figure \ref{fig:Calibration_SkinCancer}, for small-scale datasets such as chest cancer, the introduction of DA can lead to a lower median value of ECE $\sim 3\%$ compared to without DA ($\sim 4\%$). For multimodal data sets such as MC, after DA, the deep model also exhibits a lower median ECE value $\sim 6.82\%$ compared to without DA ($\sim 8.2\%)$. We argue that the use of DA adjusts the image features into distinct clusters with a clear decision boundary, thereby improving the confidence of the model predictions and resulting in a lower ECE value. These findings highlight the potential of DA to calibrate the classifier model related to MRI and CT images.

However, the use of DA does not always guarantee that the model is well calibrated. As illustrated in Figure \ref{fig:Calibration_SkinCancer}, the use of DA results in a large median value $\sim 8.5\%$ compared to no DA ($\sim 6.8\%$). This suggests that for class-unbalanced datasets such as skin cancer, simply DA may result in a negative adaptation, thus increasing its ECE value.

\section{Conclusion}\label{S8}
This paper introduced the concept of common DA techniques. Furthermore, we performed simulations of these DA approaches using CNNs with public medical data sets. The findings suggest that DA can improve model performance in the case of data fusion and brain tumor classification. Furthermore, the use of DA can improve the interpretability of the model in the classification of skin cancer. In addition, DA provides better calibration effects for CC and BT datasets. For future work, additional research is required to address the challenges faced by the class imbalance in DA for classification tasks.


{\small
\bibliographystyle{ieeetr}
\bibliography{name}
}

\end{document}